\documentclass[conference]{IEEEtran}
\IEEEoverridecommandlockouts
\usepackage{cite}
\usepackage{amsmath,amssymb,amsfonts}
\usepackage{algorithmic}
\usepackage{graphicx}
\usepackage{textcomp}
\usepackage{xcolor}
\usepackage[colorlinks]{hyperref}
\hypersetup{
     colorlinks,
     linkcolor=black,
     citecolor=black,
     filecolor=black,
     urlcolor=blue,
 }
\def\BibTeX{{\rm B\kern-.05em{\sc i\kern-.025em b}\kern-.08em
    T\kern-.1667em\lower.7ex\hbox{E}\kern-.125emX}}
\begin{document}

\title{Investigating Lagrangian Neural Networks for Infinite Horizon Planning in Quadrupedal Locomotion}

\author{
Prakrut Kotecha$^{1}$, Aditya Shirwatkar$^{1}$, Shishir Kolathaya$^{2}$
\thanks{$^{1}$P. Kotecha, A. Shirwatkar are with the Robert Bosch Center for Cyber-Physical Systems, Indian Institute of Science, Bengaluru.}%
\thanks{$^{2}$S. Kolathaya is with the Robert Bosch Center for Cyber-Physical Systems and the Department of Computer Science \& Automation, Indian Institute of Science, Bengaluru.}
\thanks{Email: \href{mailto:stochlab@iisc.ac.in}{stochlab@iisc.ac.in}}%
}

\maketitle

\begin{abstract}
Lagrangian Neural Networks (LNNs) present a principled and interpretable framework for learning the system dynamics by utilizing inductive biases. While traditional dynamics models struggle with compounding errors over long horizons, LNNs intrinsically preserve the physical laws governing any system, enabling accurate and stable predictions essential for sustainable locomotion. This work evaluates LNNs for infinite horizon planning in quadrupedal robots through four dynamics models: (1) full-order forward dynamics (FD) training and inference, (2) diagonalized representation of Mass Matrix in full order FD, (3) full-order inverse dynamics (ID) training with FD inference, (4) reduced-order modeling via torso centre-of-mass (CoM) dynamics. Experiments demonstrate that LNNs bring improvements in sample efficiency (10x) and superior prediction accuracy (up to 2-10x) compared to baseline methods. Notably, the diagonalization approach of LNNs reduces computational complexity while retaining some interpretability, enabling real-time receding horizon control. These findings highlight the advantages of LNNs in capturing the underlying structure of system dynamics in quadrupeds, leading to improved performance and efficiency in locomotion planning and control. Additionally, our approach achieves a higher control frequency than previous LNN methods, demonstrating its potential for real-world deployment on quadrupeds.
\end{abstract}


\section{Introduction}

Accurately modeling system dynamics is fundamental in control and robotics. Traditional physics-based methods, such as Lagrangian mechanics, provide interpretable and physically consistent models but can be computationally expensive and require precise system parameters \cite{Spong1989RobotDA}. This challenge is amplified in complex systems like legged robots, where multiple degrees of freedom and external disturbances complicate modeling. Data-driven approaches offer an alternative by learning dynamics from data \cite{chen2018neural}, but they often lack interpretability and struggle to generalize beyond training conditions. To address this, hybrid methods that embed physical principles into machine learning frameworks have gained interest \cite{raissi2017physicsI, raissi2017physicsII, raissi2019physics, djeumou2022neuralnetworksphysicsinformedarchitectures}, combining the strengths of both approaches to improve generalization and efficiency.

\begin{figure}
    \centering
    \includegraphics[width=0.4\textwidth]{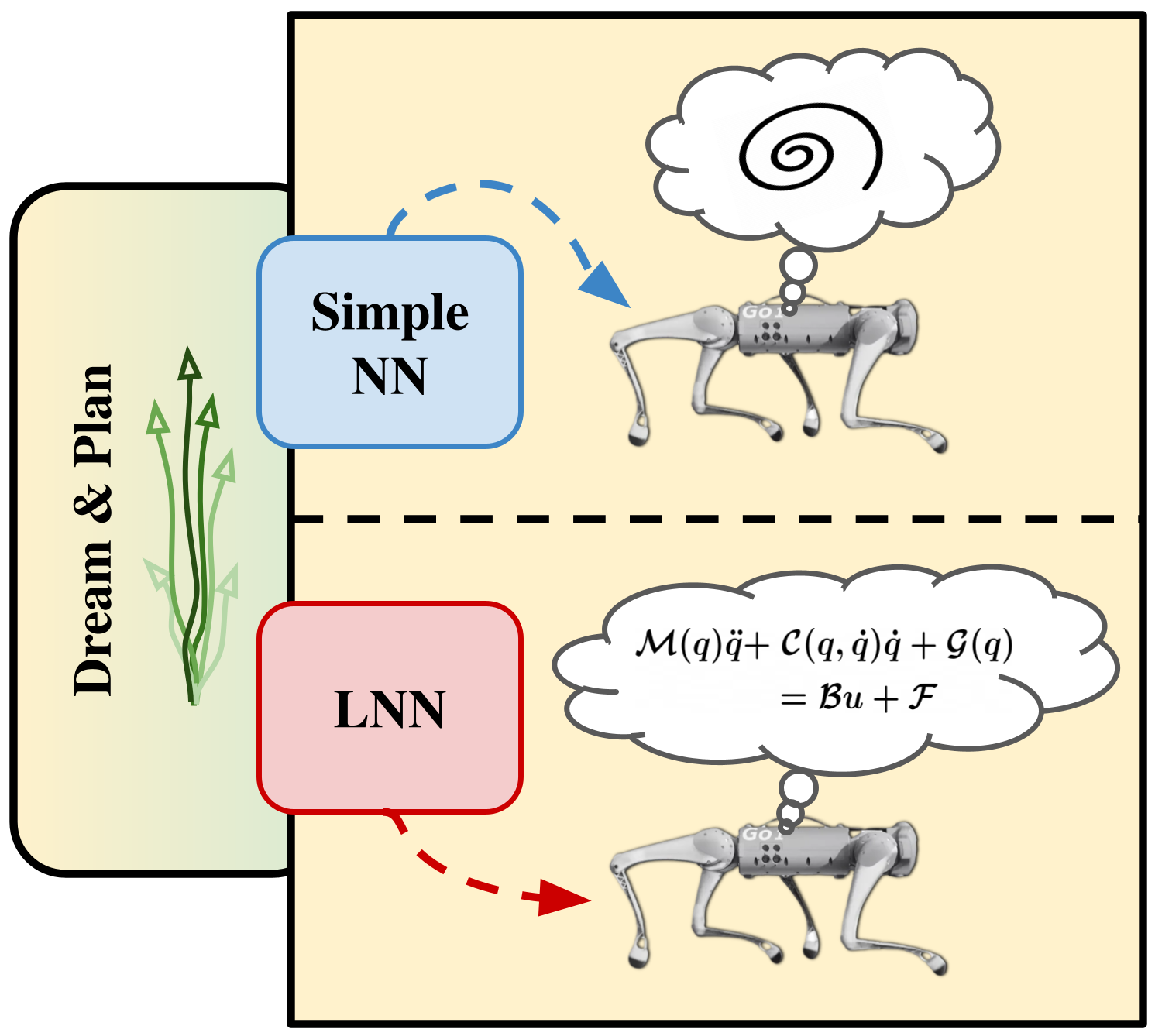}
    \caption{Traditional neural networks (Simple NN) learn dynamics in a purely data-driven manner, often resulting in an opaque representation of motion. In contrast, Lagrangian Neural Networks (LNN) embed physical principles into their design, incorporating known equations of motion. This structured approach not only improves interpretability but also ensures that the learned dynamics are more consistent with real-world physics.}
    \label{Main figure}
\end{figure}

Recent work has also explored Lagrangian Neural Networks (LNNs) \cite{LNN, cartesianLNN, gruver2022deconstructinginductivebiaseshamiltonian, finzi2020simplifyinghamiltonianlagrangianneural}, which integrate known physical principles directly into learned dynamics. Unlike purely data-driven models that learn differential equations governing the system, LNNs leverage Lagrangian mechanics to ensure physically consistent predictions, leading to better generalization, improved sample efficiency, and enhanced interpretability. This inductive bias not only improves generalization but also ensures that the learned models adhere to physical constraints, making them more reliable for control applications. Furthermore, LNNs offer a principled way to bridge the gap between traditional model-based methods and modern data-driven techniques, providing a pathway to scalable and interpretable dynamics modeling.

The application of such physics-informed learning frameworks to real-world systems, particularly in robotics, remains an area of active research. DeLan \cite{delan} was one of the first works to demonstrate the viability of LNNs in low-degree-of-freedom (DoF) systems. Legged robots, for instance, present unique challenges due to their high dimensionality, underactuation, and complex contact dynamics. Accurate modeling of these systems is critical for achieving robust and efficient locomotion, yet traditional methods often fall short in capturing the full complexity of their behavior. As the state-of-the-art continues to advance toward more agile and autonomous systems, the need for scalable, interpretable, and physically consistent dynamics models becomes increasingly pressing.

Model-free reinforcement learning (RL) combined with model predictive control (MPC) has significantly improved the performance of complex dynamical systems \cite{polo, loop, d3p, mopac, dmdmpc, tdmpc, tdmpc2, piploco}. While these approaches outperform traditional techniques, they lack an explicit understanding of system dynamics, limiting their interpretability. To address this, PIPLoco \cite{piploco} introduces a co-learning approach that integrates Dreamer and RL networks, enabling sequential dynamics learning for improved performance. This framework has been successfully tested using baseline neural networks as a dynamics model, demonstrating promising results. Figure \ref{Main figure} shows the difference in dreaming for purely data-driven dynamics model and a physics informed dynamics model in a pictorial sense.

Despite these advancements, existing implementations of Lagrangian Neural Networks (LNNs) face computational bottlenecks that hinder their direct adoption. This challenge is particularly evident in quadrupeds, where inverting the mass matrix during inference is computationally expensive within a sampling-based planner \cite{mppi}. Addressing these limitations is crucial for making LNN-based approaches more practical for real-time control in robotics.

\subsection{Contribution}

In this work, we conduct a comprehensive ablation study on Lagrangian Neural Networks (LNNs) for quadruped locomotion, analyzing their efficiency and predictive capabilities within a planning framework. Specifically, we investigate the impact of (i) inverse dynamics training for improved computational efficiency, (ii) state-space reduction via torso CoM representation for better computational efficiency, and (iii) efficient mass matrix handling via diagonalization to enhance inference speed. Our results benchmark LNNs against existing approaches, demonstrating superior accuracy and sample efficiency in modelling quadruped dynamics. By incorporating these improvements, our method combines the benefits of physics-informed planning with computational efficiency, making it a practical and scalable solution for real-world quadruped locomotion. We incorporate the PIPLoco \cite{piploco} framework to enhance our approach.

\begin{figure*}[!ht]
    \centering
    \includegraphics[width=0.8\textwidth]{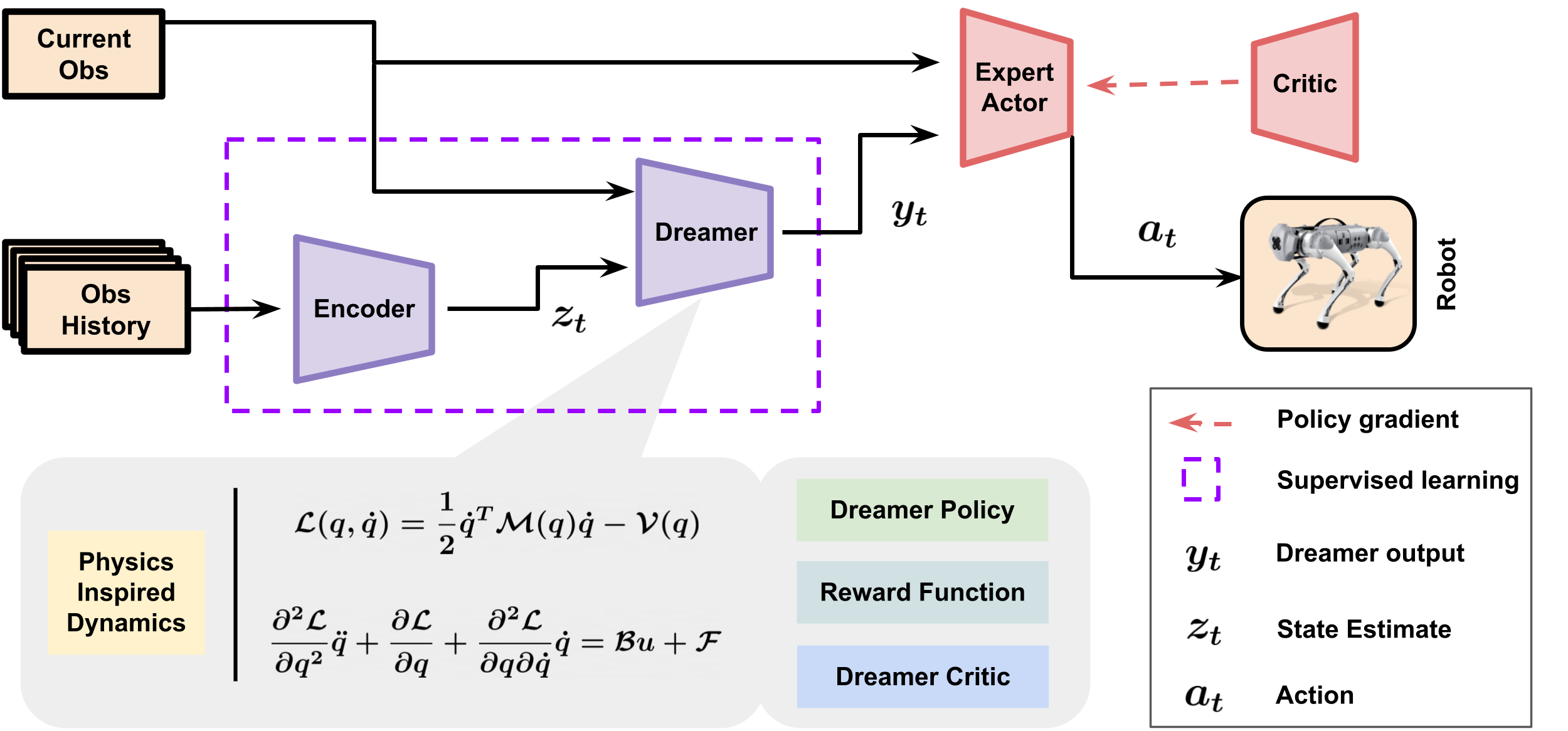}
    \caption{The framework leverages physics-inspired dynamics within a Dreamer-based model to ensure physical consistency in quadruped locomotion. The internal model, comprising a state estimator and Dreamer module, co-learns with the Asymmetric Actor-Critic to generate more accurate and physically consistent motion.}
    \label{fig:training}
\end{figure*}

\section{Setup}  
\label{sc:setup}

In this section, we describe the approach incorporated, starting with the problem formulation and then detailing the training and deployment processes.

\subsection{Preliminaries}  

Quadruped locomotion is modeled as an infinite-horizon discounted Partially Observable Markov Decision Process (POMDP), defined by the tuple:  
\begin{equation}  
\mathcal{M} = \{\mathcal{S}, \mathcal{A}, \mathcal{O}, r, \mathcal{P}, \gamma, \mathcal{F}\}
\end{equation}  
where $\mathcal{S} \subset \mathbb{R}^n$ represents the state space, $\mathcal{O} \subset \mathbb{R}^p$ the observation space, and $\mathcal{A} \subset \mathbb{R}^m$ the action space. The system dynamics are governed by the transition function $\mathcal{P}: \mathcal{S} \times \mathcal{A} \mapsto \Pr(\cdot)$, with $\Pr$ being a probability measure, while the observation function $\mathcal{F}: \mathcal{S} \mapsto \mathcal{O}$ provides partial observability. The reward function $r : \mathcal{S} \times \mathcal{A} \mapsto \mathbb{R}$ guides the learning process, and $\gamma \in (0,1)$ is the discount factor.  

To solve this POMDP, we employ an Asymmetric Actor-Critic framework \cite{asymmetric}, where the critic operates with privileged information not available to the expert actor. We additionally incorporate an architecture similar to \cite{piploco}. In our case we consider an encoder that estimates states, a Dreamer that comprises: an LNN as a dynamics model, a policy, a learnt reward  and value function. Subsequently we will discuss the details on how all these models are obtained.

\subsection{Training} \label{Training}

The training setup consists of an expert actor, a privileged critic, an encoder and the Dreamer. The expert actor and critic are trained using Proximal Policy Optimization (PPO) \cite{ppo}, while the encoder and Dreamer are learned via supervised training (see Fig.~\ref{fig:training}). 

\textbf{Observation \& Action Space:} The observations in our task includes the robot’s joint angles relative to their nominal values, joint velocities, projected gravity, and previous actions. Additionally, velocity commands for the base, including linear velocity $\text{v}^{cmd}_{xy}$ and angular velocity $\omega_z^{cmd}$, are provided. The action space consists of joint-angle perturbations applied to the nominal joint angles, where the commanded joint position at time $t$ is given by: $q_t = a_t + q_{\text{nominal}}.$  
These commands are processed through an actuator network \cite{eth2019} to compute the torques required for execution.  

\textbf{Reward:} The reward function encourages stable and agile locomotion by incorporating terms for achieving target velocity commands, maintaining body height, and stabilizing orientation. Additionally, we introduce a barrier function-based reward \cite{barrier} to enforce smooth and controlled movement, reducing abrupt or aggressive changes in motion.  

\textbf{Dreamer:} To enhance the robustness of RL policy, we introduce an approach inspired by \cite{piploco}. Our model consists of a state estimator $E_\phi$ and a Dreamer module, which enables the expert actor to anticipate future observations. Specifically, the Dreamer module includes a dynamics model $d_\theta$, an external force estimator $F_\theta$, a cloned policy $\pi_\theta$, a reward model $r_\theta$, a value function $V_\theta$. Unlike previous approaches \cite{dreamwaq, himloco}, this allows the expert actor to reason beyond reactive control and adaptively adjust its actions.

\section{Methodology}

In this section, we describe the neural network architectures used for learning dynamics models in the Dreamer for a given task. Specifically, we compare a structure-agnostic neural network with a physics-informed approach based on Deep Lagrangian Neural Networks (LNNs). The goal is to analyze the impact of incorporating explicit physical priors into the learning process.

Given an observation vector  
\begin{equation}
o = \{\Theta, \dot{\Theta}, v^{cmd}, g_p\} \in \mathcal{O},
\end{equation}
where \( \Theta \) represents joint angles, \( \dot{\Theta} \) joint velocities, \( v^{cmd} \) the command velocity, and \( g_p \) the projected gravity, the objective is to learn a predictive model that estimates future states \( o_{next} \) based on different structural assumptions.

\textbf{Baseline Neural Network approach: (BNN)}
A baseline data-driven approach is implemented using a Multi-Layer Perceptron (MLP) that maps the current observation \( o\) to the predicted next observation \( o_{next} \):
\begin{equation}
o_{next} = d_\theta(o).
\end{equation}
This model is trained via supervised learning to minimize the mean squared error (MSE) loss between predicted and ground-truth observations:
\begin{equation}
\mathcal{L}_{d_\theta} = ||o_{next} - \hat{o}_{next}||^2.
\end{equation}
Where $\hat{o}_{next}$ is the true value of the next observation taken from the environment.
Since this approach does not explicitly enforce physical consistency, it serves as an experiment to assess the benefits of physics-aware models.

\subsection{Lagrangian Neural Networks}
To integrate physical structure, we employ a LNN that predicts system dynamics using an explicit Lagrangian formulation. Instead of directly regressing future states, LNN models the underlying physics by learning the mass matrix \( M(q) \), which governs system dynamics. However, since the observation space does not provide the full system state, a state estimator network \( E_\theta \) is introduced to infer the full state from partial observations. This estimator is trained using supervised learning with an MSE loss between the predicted and true state:
\begin{subequations}
    \begin{equation}
        q = s_\theta(o).
    \end{equation}
    \begin{equation}
        \mathcal{L}_{s_\theta} = ||q_{next} - \hat{q}_{next}||^2.
    \end{equation}
\end{subequations}

Where 
$q \in \mathbb{R}^{18}$ contains the full generalized state of the robot at the current time step. 

Following \cite{delan}, the mass-inertia matrix \( M(q) \) is parameterized as a positive definite symmetric matrix:
\begin{equation}
M(q) = Y(q)Y(q)^T + \epsilon I,
\end{equation}
where \( Y(q) \) is a learnable lower triangular matrix, and \( \epsilon I \) ensures positive definiteness for the resulting matrix. This structure enforces physical consistency while maintaining flexibility in learning dynamics. Using this, we can calculate the Lagrangian of the system 
\begin{equation}
    L_\theta(q,\dot q) = \dot q^T M_\theta(q) \dot q - V_\theta(q),
\end{equation}
where \( V_\theta \) is the potential energy of the system.
From this, the LNN predicts system evolution by computing joint accelerations \( \ddot{q} \) from torques \( u \) via forward dynamics:
\begin{equation}\label{eqn:forward_lnn}
    \ddot q = \Bigg ( \frac{\partial^2 L_\theta}{\partial \dot q^2} \Bigg)^{-1} \Bigg ( \frac{\partial L_\theta}{\partial q} - \frac{\partial^2 L_\theta}{\partial q \partial \dot q} \dot q + Bu + F_\theta \Bigg).
\end{equation}
Where \(B\) is an 18x12 matrix with a 12x12 identity matrix concatenated with a 6x12 zero matrix. Since reinforcement learning policies output $a$ which can be converted to torques \( u \), this formulation aligns well with policy interaction and real-time control.

With \( \ddot q\) we can find \( [q_{next}, \dot q_{next}]\) by integration using,
\begin{subequations}
\begin{equation}
    \dot q_{next} = \dot q + \ddot q*dt
\end{equation}
\begin{equation}
    q_{next} = q + \dot q_{next}*dt
\end{equation}  
\label{integration}
\end{subequations}

\textbf{Diagonalisation of Mass matrix:} In the forward dynamics equation \eqref{eqn:forward_lnn}, there is a computational bottleneck involved. Since inverting a non-diagonal matrix is computationally expensive, we leverage the fact that \( M(q) \) is a positive definite symmetric matrix, allowing it to be diagonalised as:
\begin{equation}
    M(q) = P(q)^T \Lambda(q) P(q),
\end{equation}
where \( P(q) \) is a matrix whose columns are the eigenvectors of \( M(q) \), and \( \Lambda(q) \) is a diagonal matrix containing the eigenvalues. Since \( M(q) \) is symmetric, its eigenvectors form an orthonormal basis, allowing the inverse to be computed efficiently as:
\begin{equation}
    M(q)^{-1} = P(q) \Lambda(q)^{-1} P(q)^T.
\end{equation}
Since \( \Lambda(q) \) is diagonal, its inverse is simply \( \Lambda(q)^{-1} = \text{diag}(1/\lambda_i) \), where \( \lambda_i \) are the eigenvalues. 
Further, we can simplify equation \eqref{eqn:forward_lnn} to obtain the following form:
\begin{equation}
M(q) \ddot{q} + C(q, \dot{q}) \dot{q} + G(q) = B u + F_{ext}.
\label{eq:newton_euler}
\end{equation}
Here $C$ is commonly known as the Coriolis Matrix and $G$ is the Bias vector. This can be finally rewritten as:
\begin{equation}
    P(q)^T \Lambda(q) P(q) \ddot{q} + C(q, \dot{q}) \dot{q} + G(q) = B u + F_{ext}.
\end{equation}
Solving for \( \ddot{q} \) simplifies to:
\begin{equation}
    \ddot{q} = P(q) \Lambda(q)^{-1} P(q)^T \left(- C(q, \dot{q}) \dot{q} - G(q) + B u + F_{ext} \right).
\end{equation}
This reduces computational complexity and enables faster inference, which is crucial for real-time control.

\textbf{Training with inverse dynamics:}
An alternative training strategy is considered where the model learns inverse dynamics instead of forward dynamics. In this approach, the model is trained to predict the required torques \( Bu \) given observed accelerations:
\begin{equation}
Bu = \Bigg( \frac{\partial^2 L_\theta}{\partial \dot q^2} \Bigg) \ddot q - \Bigg( \frac{\partial L_\theta}{\partial q} \Bigg) + \Bigg( \frac{\partial^2 L_\theta}{\partial q \partial \dot q} \Bigg) \dot q - F_\theta.
\end{equation}
Training in this way avoids numerical instabilities and computation associated with matrix inversion. During inference, the model operates in a forward dynamics setting to maintain consistency with RL policy execution. Since the acceleration $\ddot q$ is unknown during prediction, inverse dynamics cannot be directly applied. Instead, we use forward dynamics to compute $\ddot q$ from the policy-generated torques $u$ (converting $a$ to $u$ using actuator net \cite{eth2019}) and then integrate it using Equation \ref{integration} to obtain the next state.

\textbf{Training with Center of Mass (CoM) state of torso only:}
An explicit representation of the torso’s center of mass (CoM) is incorporated to make it further computationally efficient. By using CoM-based kinematics rather than raw joint-space observations, the input dimensionality is reduced while preserving essential motion characteristics. This structured representation improves generalization and stability.

\begin{equation}
    \ddot X = \Bigg ( \frac{\partial^2 L_\theta}{\partial \dot X^2} \Bigg)^{-1} \Bigg ( - \frac{\partial L_\theta}{\partial X} - \frac{\partial^2 L_\theta}{\partial X \partial \dot X} + F_\theta \Bigg).
\end{equation}
Where, 
$X \in \mathbb{R}^{12}$ consists of the pose and twist of the CoM. Here, due to a reduction in dimensionality, the size of the mass matrix reduces, leading to a reduction in computation time. Although this is an approximation, it leads to a higher prediction error, but it is still a better alternative than the BNN method, as we will see in section \ref{results}.

\subsection{Deployment Scheme}
\label{sec:deployment}

To deploy our learned policies effectively, we integrate a model-based planning strategy leveraging the \textit{PIP-Loco} framework \cite{piploco}. Unlike traditional model-free reinforcement learning (RL) methods that execute policies reactively, PIP-Loco incorporates dreamer module to anticipate future observations and optimize control actions over an infinite horizon in an \textit{online} fashion. This approach improves robustness, constraint satisfaction, and adaptability to complex terrains.

Since we will be writing an optimisation formulation we will be using all the symbols defined with ${symbol}_t$ for time step information. Also here we are using actions $a_t$ as defined in section \ref{Training}. 

We define the optimal action sequence $a^*_{t:t+H}$ at observation $o_t$ by solving the constrained optimization problem:

\begin{equation}
    \max_{(a_t, \dots, a_{t+H})} \sum_{k=0}^{H-1} \gamma^k r_{\theta}(q_{t+k}, a_{t+k}) + \gamma^H V_{\theta}(q_{t+H})
\end{equation}
\begin{equation}
    \text{s.t. } q_{t+k} = E_\theta(o_{t+k}), \quad q_{t+k+1} = d_{\theta}(q_{t+k}, a_{t+k}),
\end{equation}

Where $\gamma$ is the discount factor, $V_{\theta}$ is the learned value function, $d_\theta$ is the learned dynamics model, and $E_\theta$ gives the full state ($q$) information.

\textbf{Integration with LNNs:}
To enhance computational efficiency and physical consistency, we integrate our proposed approaches within the PIP-Loco framework. Instead of directly regressing future states, the LNN enforces structured physics-based constraints. Our learned forward dynamics model is used in place of $d_\theta$ using equation \ref{eqn:forward_lnn} \& equation \ref{eq:newton_euler}, to predict motion trajectories for planning. The LNN-based forward dynamics is particularly beneficial in scenarios requiring long-horizon stability. This can be a promising avenue to look for in future work.

\section{Results}
\label{results}

This section details the training and deployment of our reinforcement learning (RL) framework and Dreamer models. We compare various dynamics models using prediction error and dynamics loss graphs, demonstrating improved prediction accuracy and greater sample efficiency. 

For training, we utilized an open-source environment for the Unitree Go1 quadruped robot, based on Nvidia's Isaac Gym simulator \cite{isaac_gym, leggedgym}. The neural networks for PPO were implemented using the PyTorch framework \cite{pytorch}, while the dreamer was implemented using JAX \cite{jax}. The training and inference of this method was conducted on a desktop system equipped with an Intel Xeon(R) Gold 5318Y CPU (48 cores, 2.10 GHz), 512 GB of RAM, and an NVIDIA RTX 6000 Ada Generation GPU.

\subsection{Implementation Details}

\subsubsection{Training}
In the RL framework, we employ an actor-critic architecture with hidden layers of sizes [512, 256, 128] implemented as a multi-layer perceptron (MLP) with \texttt{tanh} activations. Similarly, for the encoder and Dreamer, we use hidden layers of size [256, 256] with \texttt{tanh} activations. Since these networks are trained in a supervised fashion, a smaller architecture does not negatively impact performance.

During training, privileged information about the next state is stored in the buffer, enabling the training of the dynamics model. The base RL algorithm is PPO\cite{ppo} with certain modifications tailored for our quadruped environment \cite{himloco}.

\subsubsection{Deployment}
During deployment, the Dreamer, implemented in JAX \cite{jax}, provides policy actions along with future state predictions. The dreamer policy is trained in a supervised manner using actions generated by the expert RL policy. This enables effective planning and can be integrated into warm-starting Model Predictive Control (MPC) for hardware deployment, reducing inference time.

Since privileged information is unavailable during deployment, inverse dynamics cannot be directly applied as it requires the current joint acceleration (\(\ddot{q}\)) for torque calculation. Instead, our policy outputs desired joint angles (\(a\)), which are converted into joint torques (\(\tau\)) using an actuator network \cite{eth2019}. This limitation will be addressed in future work.

\begin{figure}[!ht]
    \centering
    \includegraphics[width=0.48\textwidth]{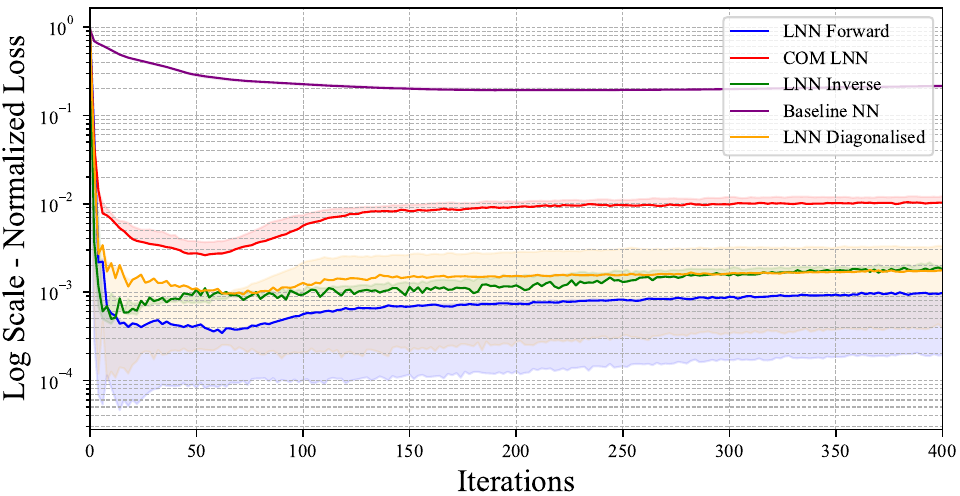}
    \caption{Dynamics loss graph plotted on a logarithmic scale for better comparison.}
    \label{fig:dynamics_loss}
\end{figure}

\subsection*{Sample Efficiency}
From Figure~\ref{fig:dynamics_loss}, we observe that structured dynamics models require fewer iterations (i.e., fewer samples) to learn dynamics compared to structure-agnostic approaches. The graph is plotted on a logarithmic scale to highlight differences in methods incorporating Lagrangian dynamics. Since the initial loss values vary significantly across approaches, we normalize them to ensure consistent scaling and clearer interpretation.

Previous work \cite{delan} has shown that physics-informed learning benefits simple systems, but scaling up Lagrangian Neural Networks (LNNs) is challenging due to an increase in learnable parameters. Our results extend this observation to a more complex systems i.e. quadrupeds. While the base LNN achieves the lowest dynamics loss, all variations of LNNs ultimately show an improved sample efficieny when compared to Baseline Neural Networks (BNNs) approach.

\begin{figure}[!ht]
    \centering
    \includegraphics[width=0.48\textwidth]{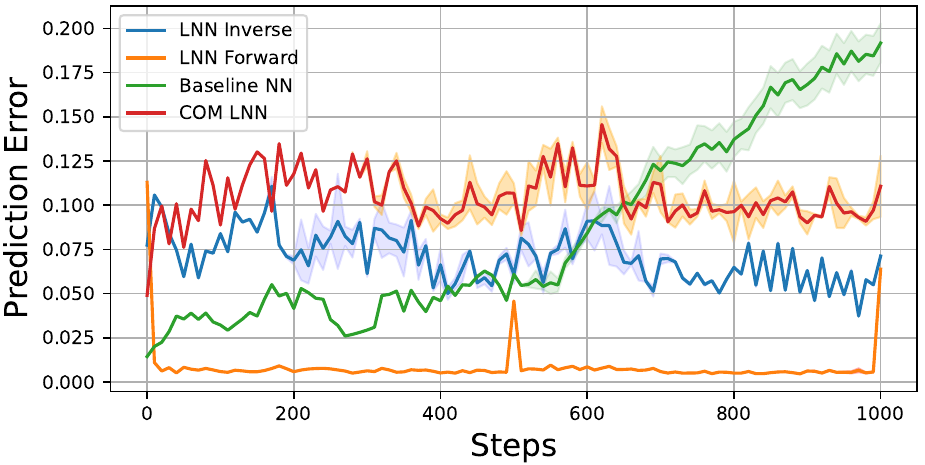}
    \caption{Prediction error of different dynamics models compared to actual state values.}
    \label{fig:prediction_error}
\end{figure}

\subsection*{Prediction Error}
Beyond faster learning, accurate dynamics prediction is crucial for deployment. Figure~\ref{fig:prediction_error} compares prediction errors against actual environment states. As we can see when compared to the BNN method the prediction error in forward LNN is lesser. On average, we see upto 85 \% reduction in the value of prediction error. The forward LNN exhibits the lowest error, closely followed by the LNN trained using inverse dynamics. This suggests that training LNNs with inverse dynamics not only accelerates computation but also maintains prediction accuracy within a conservative range. As the COM-based LNN does not capture the full essence of the system it is expected to have higher prediction error and we can see that it is comparable to the BNN method. 

This observation opens avenues for incorporating inverse dynamics into dynamics prediction modules, potentially improving training and inference time.

\begin{figure}[!ht]
    \centering
    \includegraphics[width=0.48\textwidth]{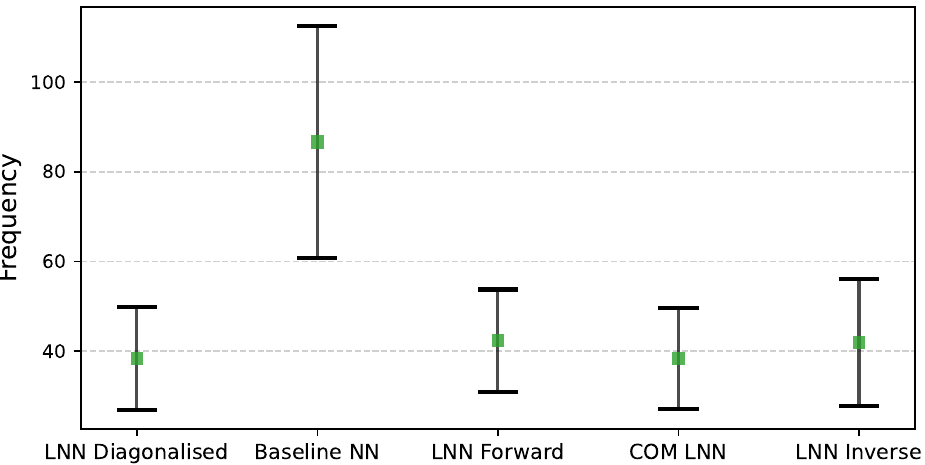}
    \caption{Controller inference frequency variation of different dynamics models.}
    \label{fig:Frequency}
\end{figure}

\subsection*{Inference Speed Analysis}
Additionally, we analyze our controller frequency across methods using candle plots (figure \ref{fig:Frequency}), revealing that physics-inspired approaches tend to have lower inference frequencies, leading to slower control loops. But also we see that inverse dynamics has lead to a higher max value of the reachable inference frequency. This motivates further optimization of our framework to enhance real-time deployment performance. Using these dynamics models in MPC for deployment might enhance the inference frequency further which can make them suitable for real-world deployment. 

\section*{Conclusion}
We compare a Baseline Neural Network (BNN) approach with a physics-informed method, emphasizing the benefits of incorporating structural priors. Our results demonstrate that physics-aware Lagrangian Neural Networks (LNNs) enhance generalization, enforce physical consistency, and integrate effectively with reinforcement learning-based control, making them a promising choice for modeling robot dynamics. Additionally, we validate the feasibility of LNNs for complex systems such as quadrupedal robots, highlighting their potential for hardware deployment with appropriate adaptations. To address computational challenges, we propose alternative approaches for implementing LNNs that improve efficiency while preserving their physical interpretability. However, the increasing number of parameters with system complexity remains a key limitation. A promising direction for future research is the incorporation of inverse dynamics to accelerate training, alongside modifications to reinforcement learning policies to output joint accelerations or directly optimize joint angles. These advancements could facilitate the deployment of physics-informed learning models in real-world robotic applications.

\bibliographystyle{ieeetr}
\bibliography{references}

\begin{thebibliography}{10}

\bibitem{Spong1989RobotDA}
M.~W. Spong, ``Robot dynamics and control,'' 1989.

\bibitem{chen2018neural}
R.~T. Chen, Y.~Rubanova, J.~Bettencourt, and D.~K. Duvenaud, ``Neural ordinary differential equations,'' {\em Advances in neural information processing systems}, vol.~31, 2018.

\bibitem{raissi2017physicsI}
M.~Raissi, P.~Perdikaris, and G.~E. Karniadakis, ``Physics informed deep learning (part i): Data-driven solutions of nonlinear partial differential equations,'' {\em arXiv preprint arXiv:1711.10561}, 2017.

\bibitem{raissi2017physicsII}
M.~Raissi, P.~Perdikaris, and G.~E. Karniadakis, ``Physics informed deep learning (part ii): Data-driven discovery of nonlinear partial differential equations,'' {\em arXiv preprint arXiv:1711.10566}, 2017.

\bibitem{raissi2019physics}
M.~Raissi, P.~Perdikaris, and G.~E. Karniadakis, ``Physics-informed neural networks: A deep learning framework for solving forward and inverse problems involving nonlinear partial differential equations,'' {\em Journal of Computational Physics}, vol.~378, pp.~686--707, 2019.

\bibitem{djeumou2022neuralnetworksphysicsinformedarchitectures}
F.~Djeumou, C.~Neary, E.~Goubault, S.~Putot, and U.~Topcu, ``Neural networks with physics-informed architectures and constraints for dynamical systems modeling,'' 2022.

\bibitem{LNN}
M.~Cranmer, S.~Greydanus, S.~Hoyer, P.~Battaglia, D.~Spergel, and S.~Ho, ``Lagrangian neural networks,'' {\em arXiv preprint arXiv:2003.04630}, 2020.

\bibitem{cartesianLNN}
M.~Finzi, K.~A. Wang, and A.~G. Wilson, ``Simplifying hamiltonian and lagrangian neural networks via explicit constraints,'' {\em Advances in neural information processing systems}, vol.~33, pp.~13880--13889, 2020.

\bibitem{gruver2022deconstructinginductivebiaseshamiltonian}
N.~Gruver, M.~Finzi, S.~Stanton, and A.~G. Wilson, ``Deconstructing the inductive biases of hamiltonian neural networks,'' 2022.

\bibitem{finzi2020simplifyinghamiltonianlagrangianneural}
M.~Finzi, K.~A. Wang, and A.~G. Wilson, ``Simplifying hamiltonian and lagrangian neural networks via explicit constraints,'' 2020.

\bibitem{delan}
M.~Lutter, C.~Ritter, and J.~Peters, ``Deep lagrangian networks: Using physics as model prior for deep learning,'' in {\em International Conference on Learning Representations}, 2019.

\bibitem{polo}
K.~Lowrey, A.~Rajeswaran, S.~Kakade, E.~Todorov, and I.~Mordatch, ``Plan online, learn offline: Efficient learning and exploration via model-based control,'' {\em arXiv preprint arXiv:1811.01848}, 2018.

\bibitem{loop}
H.~Sikchi, W.~Zhou, and D.~Held, ``Learning off-policy with online planning,'' in {\em Conference on Robot Learning}, pp.~1622--1633, PMLR, 2022.

\bibitem{d3p}
J.~Zhu, Y.~Wang, L.~Wu, T.~Qin, W.~gang Zhou, T.-Y. Liu, and H.~Li, ``Making better decision by directly planning in continuous control,'' in {\em International Conference on Learning Representations}, 2023.

\bibitem{mopac}
A.~S. Morgan, D.~Nandha, G.~Chalvatzaki, C.~D’Eramo, A.~M. Dollar, and J.~Peters, ``Model predictive actor-critic: Accelerating robot skill acquisition with deep reinforcement learning,'' in {\em 2021 IEEE International Conference on Robotics and Automation (ICRA)}, pp.~6672--6678, IEEE, 2021.

\bibitem{dmdmpc}
U.~A. Mishra, S.~R. Samineni, P.~Goel, C.~Kunjeti, H.~Lodha, A.~Singh, A.~Sagi, S.~Bhatnagar, and S.~Kolathaya, ``Dynamic mirror descent based model predictive control for accelerating robot learning,'' in {\em 2022 International Conference on Robotics and Automation (ICRA)}, pp.~1631--1637, IEEE, 2022.

\bibitem{tdmpc}
N.~Hansen, X.~Wang, and H.~Su, ``Temporal difference learning for model predictive control,'' {\em arXiv preprint arXiv:2203.04955}, 2022.

\bibitem{tdmpc2}
N.~Hansen, H.~Su, and X.~Wang, ``Td-mpc2: Scalable, robust world models for continuous control,'' {\em arXiv preprint arXiv:2310.16828}, 2023.

\bibitem{piploco}
A.~Shirwatkar, N.~Saxena, K.~Chandra, and S.~Kolathaya, ``Pip-loco: A proprioceptive infinite horizon planning framework for quadrupedal robot locomotion,'' 2024.

\bibitem{mppi}
G.~Williams, A.~Aldrich, and E.~Theodorou, ``Model predictive path integral control using covariance variable importance sampling,'' 2015.

\bibitem{asymmetric}
L.~Pinto, M.~Andrychowicz, P.~Welinder, W.~Zaremba, and P.~Abbeel, ``Asymmetric actor critic for image-based robot learning,'' {\em arXiv preprint arXiv:1710.06542}, 2017.

\bibitem{ppo}
J.~Schulman, F.~Wolski, P.~Dhariwal, A.~Radford, and O.~Klimov, ``Proximal policy optimization algorithms,'' {\em arXiv preprint arXiv:1707.06347}, 2017.

\bibitem{eth2019}
J.~Hwangbo, J.~Lee, A.~Dosovitskiy, D.~Bellicoso, V.~Tsounis, V.~Koltun, and M.~Hutter, ``Learning agile and dynamic motor skills for legged robots,'' {\em Science Robotics}, vol.~4, no.~26, p.~eaau5872, 2019.

\bibitem{barrier}
Nilaksh, A.~Ranjan, S.~Agrawal, A.~Jain, P.~Jagtap, and S.~Kolathaya, ``Barrier functions inspired reward shaping for reinforcement learning,'' in {\em 2024 IEEE International Conference on Robotics and Automation (ICRA)}, pp.~10807--10813, 2024.

\bibitem{dreamwaq}
I.~M.~A. Nahrendra, B.~Yu, and H.~Myung, ``Dreamwaq: Learning robust quadrupedal locomotion with implicit terrain imagination via deep reinforcement learning,'' in {\em 2023 IEEE International Conference on Robotics and Automation (ICRA)}, pp.~5078--5084, IEEE, 2023.

\bibitem{himloco}
J.~Long, Z.~Wang, Q.~Li, L.~Cao, J.~Gao, and J.~Pang, ``Hybrid internal model: Learning agile legged locomotion with simulated robot response,'' in {\em The Twelfth International Conference on Learning Representations}, 2024.

\bibitem{isaac_gym}
V.~Makoviychuk, L.~Wawrzyniak, Y.~Guo, M.~Lu, K.~Storey, M.~Macklin, D.~Hoeller, N.~Rudin, A.~Allshire, A.~Handa, and G.~State, ``Isaac gym: High performance gpu-based physics simulation for robot learning,'' 2021.

\bibitem{leggedgym}
N.~Rudin, D.~Hoeller, P.~Reist, and M.~Hutter, ``Learning to walk in minutes using massively parallel deep reinforcement learning,'' in {\em Conference on Robot Learning}, pp.~91--100, PMLR, 2022.

\bibitem{pytorch}
A.~Paszke, S.~Gross, S.~Chintala, G.~Chanan, E.~Yang, Z.~DeVito, Z.~Lin, A.~Desmaison, L.~Antiga, and A.~Lerer, ``Automatic differentiation in pytorch,'' 2017.

\bibitem{jax}
J.~Bradbury, R.~Frostig, P.~Hawkins, M.~J. Johnson, C.~Leary, D.~Maclaurin, G.~Necula, A.~Paszke, J.~Vander{P}las, S.~Wanderman-{M}ilne, and Q.~Zhang, ``{JAX}: composable transformations of {P}ython+{N}um{P}y programs,'' 2018.

\end{thebibliography}

\end{document}